\documentclass[12pt]{article}

\usepackage[margin = 1.375 in]{geometry}
\usepackage{apacite}
\usepackage{color}
\usepackage{amsmath}
\usepackage{textcomp}
\usepackage{epstopdf}
\usepackage{graphicx}
\usepackage{algorithm2e} 
\usepackage{setspace, caption} 
\numberwithin{equation}{section}

\def\W{\mbox{\bf W}}
\def\Q{\mbox{{\bf Q}}}
\def\R{\mbox{{\bf R}}}
\def\X{\mbox{{\bf X}}}
\def\Y{\mbox{$\bf{Y}$}}
\def\S{\mbox{$\bf{S}$}}
\def\s{\mbox{$\bf{s}$}}

\begin{document}

\graphicspath{{Images/}}

\bigskip
\bigskip
\begin{center}


{\bf {\Large Machine Learning of Time Series\\
Using Time-delay Embedding and Precision Annealing}}
\bigskip

    Alexander J. A. Ty, Zheng Fang, Rivver A. Gonzalez, \\

\bigskip

    Department of Physics\\
    University of California, San Diego\\
    9500 Gilman Drive\\
    La Jolla, CA   92093-0357\\

\bigskip

    Paul. J. Rozdeba,\\

\bigskip

    Institut f\"ur Mathematik\\
    Universit\"at Potsdam,\\
    Karl-Liebknecht-Str. 24-25, 14476 Potsdam, Germany\\

\bigskip

    and\\

\bigskip

    Henry D. I. Abarbanel\\

\bigskip

    Department of Physics\\
    and\\
    Marine Physical Laboratory (Scripps Institution of Oceanography)\\
    University of California, San Diego\\
    9500 Gilman Drive\\
    La Jolla, CA   92093-0357\\

\bigskip

    \today\\

\bigskip 
\end{center}


\newpage


\section*{Abstract}

    Tasking machine learning to predict segments of a time series requires estimating the parameters of a ML model with input/output pairs from the time series. We borrow two techniques used in statistical data assimilation in order to accomplish this task: (1) time-delay embedding to prepare our input data, and (2) precision annealing as a training method. The precision annealing approach identifies the global minimum of the action ($-\log[P]$). In this way we are able to identify the number of training pairs required to produce good generalizations (predictions) for the time series. We proceed from a scalar time series $s(t_n); t_n = t_0 + n \Delta t$ and using methods of nonlinear time series analysis show how to produce a $D_E > 1$ dimensional time delay embedding space in which the time series has no false neighbors as does the observed $s(t_n)$ time series. In that $D_E$-dimensional space we explore the use of feed forward multi-layer perceptrons as network models operating on $D_E$-dimensional input and producing $D_E$-dimensional outputs.  

\newpage


\section{Background}

Machine learning methods for capturing the structure of a time series with the goal of predicting future segments of that time series have been analyzed for many 
years ~\cite{Frank2001,kajitani05,goodfellow16} . We revisit this problem using analysis tools allowing one to explore questions such as: if we are given a time series data set and a network architecture with which to predict a future segment of the time series, how many distinct samples of input/output pairs used in training the network are required to achieve very good prediction (generalization)? Ascertaining the number of training examples in order to attain a given performance metric, classification error for example, have been limited to the study of learning curves in the current ML literature.

In~\cite{abar18} two of the present authors recognized for the first time the equivalence between supervised machine learning (ML) and statistical data assimilation (SDA) as widely utilized in large Physics, Geophysical, and Biophysical modeling. This recognition opens up a variety of opportunities to use methods from SDA in tasks asked of ML with the possibility of both improving the performance of ML solutions as well as gaining insight as to how these solutions work. This paper builds on~\cite{abar18} using our knowledge of how variational principles in ML may be implemented using methods not often practiced in that literature. Another of the insights in~\cite{abar18} that we called `deepest learning' when the number of layers in ML networks becomes continuous will be further visited in future publications~\cite{abar19}.

We work within a setting where we are presented with scalar time series data $s(t_n) = s(t_0 + n \Delta t) = s(n)$, sampled every $\Delta t$. A sample of these data is shown in Fig. (\ref{scalardata}). We wish to present segments of these data to a multi-layer perceptron network and train the network to learn subsequent segments of the time series. The task asked of the network in this paper is to predict one step forward in time, namely $s(n+1)$ given $s(n)$. There is no barrier to training this class of network to answer other questions about the data series. To train the selected network using the given data, we use a precision annealing (PA) method~\cite{ye2014precision,ye2015physrev}.

In this paper we explore the ability of a feed forward multi-layer perceptron (MLP) to accomplish learning this task. We show how PA allows us to answer questions about how many input/output pairs are required to achieve good generalization, namely, allowing the trained network to reliably predict from inputs not seen in the network training phase.  In cases where there are practical limitations to the number of training data are available, (e.g. cost, ethical considerations, rarity, etc.) it is of interest to determine this. Our networks have only a few hidden layers, though there seems to be no barrier to making the network much deeper. The method we present can be used with other network architectures, for example, recurrent networks, with no fundamental change in approach. We address this configuration in the later parts of this paper.


\section{Preparing the Data}

\begin{figure}[tbph] 
  \centering
  \includegraphics[width=5.67in,height=4.34in,keepaspectratio]{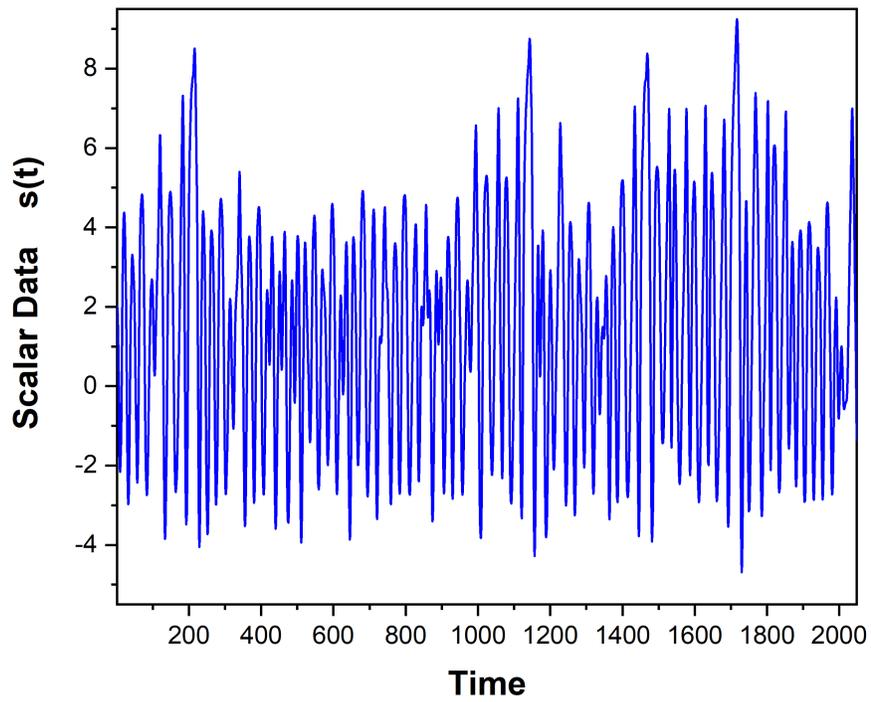}
  \caption{2048 samples of the scalar data $s(t_n = t_0 + n \Delta t) = s(n)$ that comprise our data.}
  \label{scalardata}
\end{figure}

We are presented with a time series, part of which is shown in Fig. (\ref{scalardata}). The data set is comprised of a large number of data values uniformly sampled in time at times $t_n = t_0 + n \Delta t$. We do not know $\Delta t$. We are not given any further information about the sequence $\{s(n);\;n=0,1,...N\}$. We were given $\approx 10^5$ data points, and we discarded about 
10$^{4}$ of them to eliminate potential `transients'. Only 2048 of the data points are shown in Fig. (\ref{scalardata}). 

Our goal here is to train a feedforward MLP network architecture to give as output $s(n+1)$ when presented with input $s(n)$. We could have used the method described here to train the network to predict $s(n+K)$ for any integer $K \ge 1$; we restrict our discussion here to $K = 1$. For larger values of $K$, extra caution would be required to assure that $K$ is not so large that the input and output are not correlated.

Without further knowledge of the signal $s(n)$, we assume that although it is a sequence of scalars, it might have come from projection onto the $s$-axis from the operation of a higher dimensional dynamical system. To examine this we seek a `proxy space' which carries the essential properties of the original higher dimensional source of the observed signal $s(n)$. For this purpose we turn to techniques of nonlinear time series analysis~\cite{abar96,kantz04}.


\section{Time-Delay Vectors}

If the observed time series $s(n)$ comes from projecting onto the $s$-axis, then points which appear to be nearby in time $t_n = t_0 + n \Delta t$ may be neighbors due to the projection rather than due to the dynamics that moves the actual system of interest forward in time in a higher dimensional space. Nonlinear time series methods for unfolding the scalar time series~\cite{aeyels81a,aeyels81b,takens81} use the data $s(t_n) = s(n)$ along with the time delays of the data at time points $t_n + q \tau  \Delta t = t_0 + (n + q \tau)\Delta t:\;s(n+ q \tau)$. $\tau$ and $q$ are integers.

The idea here is that $s(n+\tau)$ contains information on how the dynamics of the source of the time series $s(n) \to s(n+\tau)$ moves the system of interest forward in time. This information is not available in $s(n)$ alone.

This leads us to form the $D_E$-dimensional vector extending from each time $t_n$:
\begin{equation}
    \s(n)=[s(n),s(n+\tau),s(n+2\tau),...,s(n+ (D_E - 1)\tau)],
\end{equation} 
or, in components,
\begin{equation}
    S_q(n) = s(n + (q-1)\tau);\;q=1,2,...,D_E;
\end{equation}
$D_E$ is also an integer.

To use this idea in a practical sense, we must estimate the time delay $\tau$ and the dimension $D_E$ of the vectors $\S(n)$ containing the properties of the original state space from which $s(n)$ is projected. The value of $\tau$ should not be too small or the system will not have revealed the new information coming from the operation of the underlying dynamics, and $\tau$ should also not be too large or noise and intrinsic instabilities of the (nonlinear) dynamics will erase the utility of information at time $n + (D_E - 1)\tau$.

\subsection{Selecting $\tau$}
To estimate $\tau$ we use an information theoretic `correlation function', the average mutual information (AMI)~\cite{fano,fraser86}. This function is nonnegative~\cite{fano}, and the Fraser~\cite{fraser86} criterion is to select the first minimum of the AMI as a balance between $\tau$ being too large or too small. The minimum means that the coordinates $s(n)$ and $s(n + \tau)$ are correlated, but not so strongly correlated that no new information on the origin of the time series results from knowing both $s(n)$ and $s(n + \tau)$. 

\begin{figure}[tbph] 
  \centering
      \includegraphics[trim={300 100 400 100},clip,width=.48\textwidth]{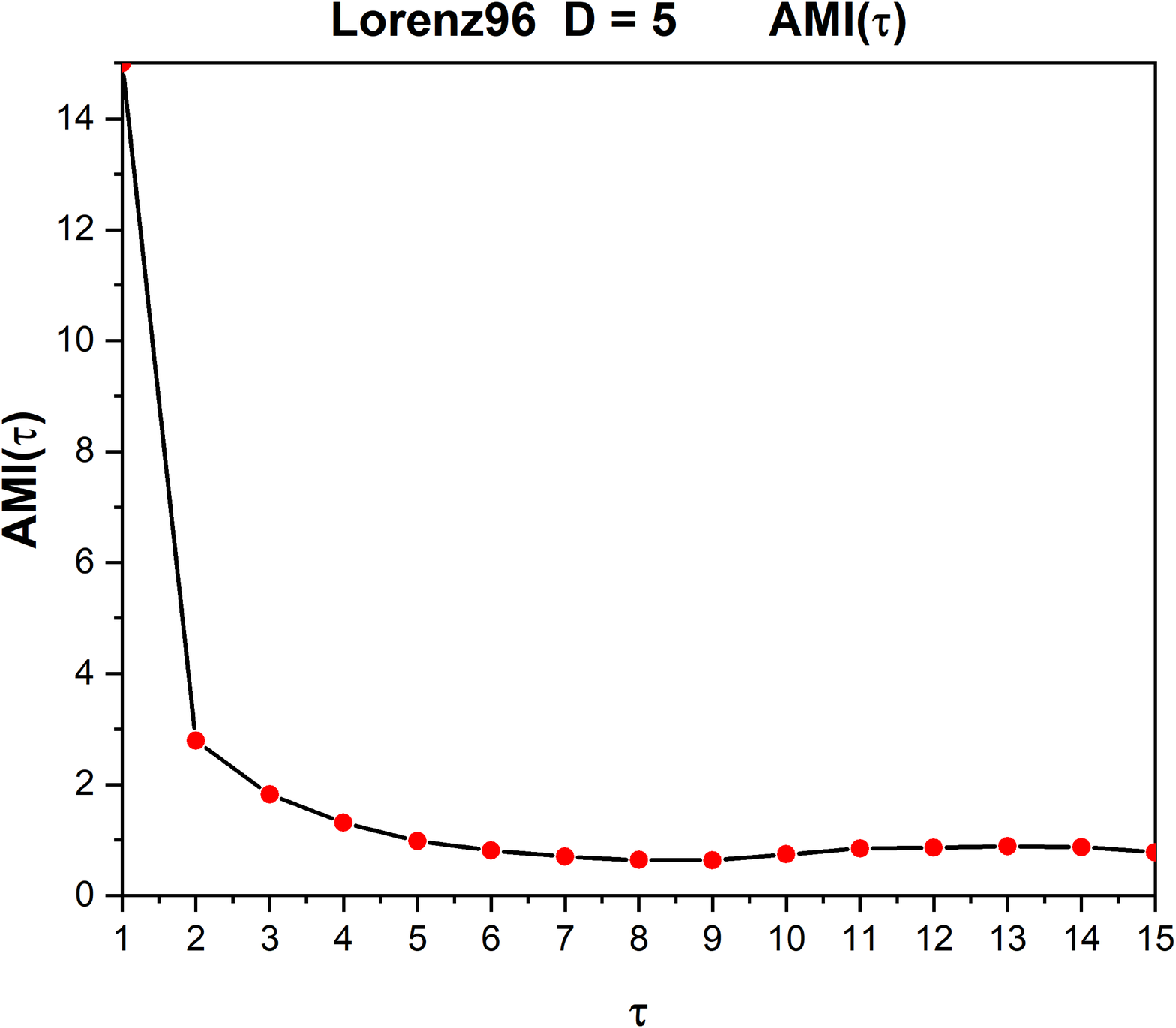}
      \includegraphics[trim={300 100 400 100},clip,width=0.48\textwidth]{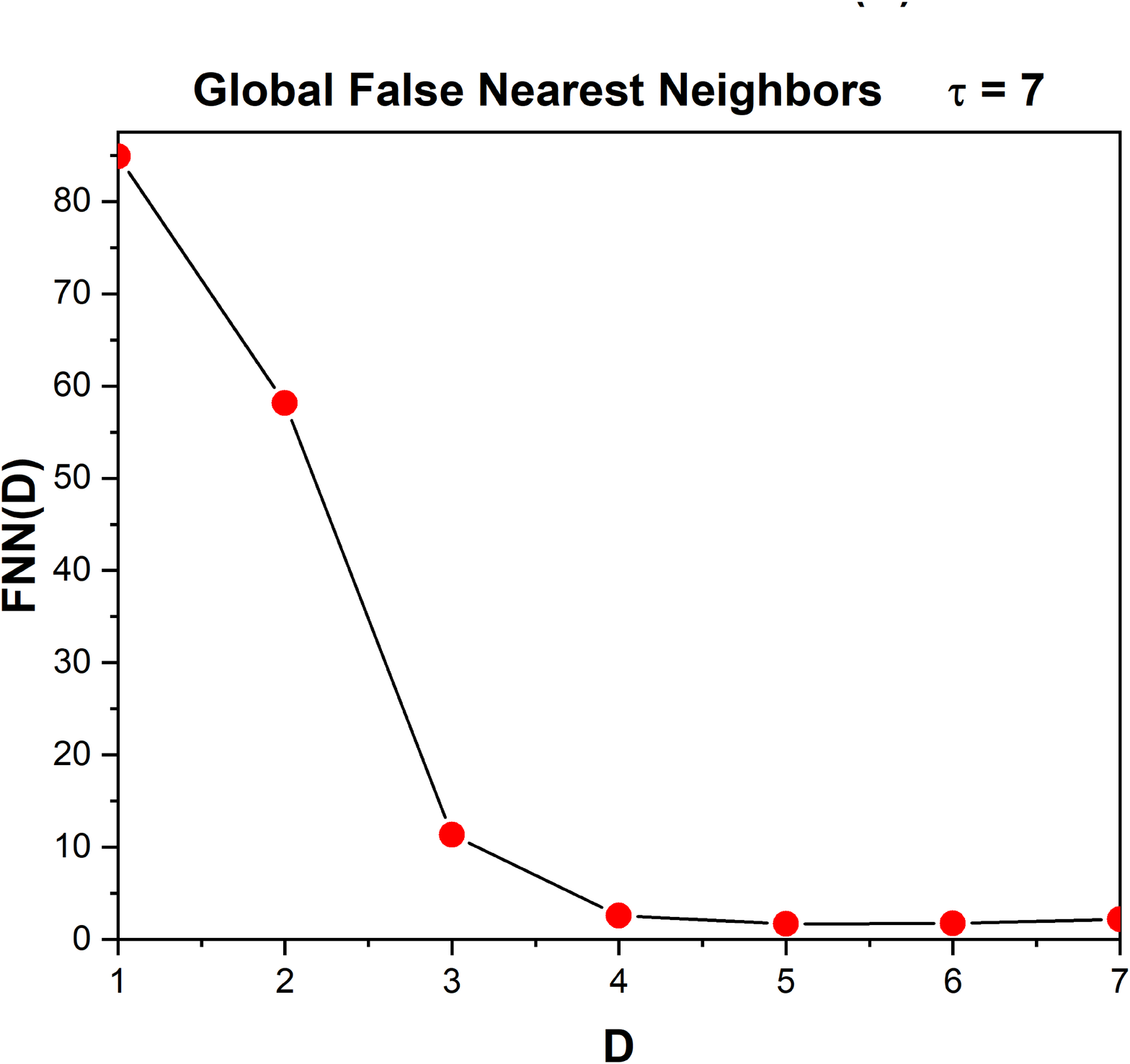}
  \caption{\textbf{Left:} Average Mutual Information Eq. (\protect\ref{amieqn}), $AMI(\tau)$ between $s(n)$ and $s(n + \tau)$ as a function of the time delay $\tau$. Using the Fraser~\protect\cite{fraser86} criterion, we select the first minimum near $\tau \approx 7-8$ for a useful time delay. \textbf{Right:} The false nearest neighbor criterion for selecting a global embedding dimension $D_E$ for vectors whose components are separated by the time delay $\tau$ using $AMI(\tau)$.}
\label{lor96d5amidt05}
\end{figure}

$AMI(\tau)$ requires the joint distribution of $\{s(n),s(n+\tau)\}, \; P(s(n),s(n+\tau))$, as well as $P(s(n))$ and $P(s(n+\tau))$. The latter come from the marginal distributions of $P(s(n),s(n+\tau))$.
\begin{equation}
    AMI(\tau) = \sum_{\{s(n),s(n+\tau)\}} P(s(n),s(n+\tau)) \log\biggl[\frac{P(s(n),s(n+\tau))}{P(s(n))\,P(s(n+\tau))}\biggr].
\label{amieqn}
\end{equation} 
This quantity, $AMI(\tau)$,  answers the question: how much information (in bits if the logarithm is to base 2) do we learn from $s(n)$ about $s(n+\tau)$ on the average over all joint values of $\{s(n),s(n+\tau)\}$.

In the {Left Panel} of Fig. (\ref{lor96d5amidt05}) we display  $AMI(\tau)$ evaluated from 2$^{15}$ samples of the time series of $s(n)$. The first minimum of this is near $\tau \approx\,$ 7 or 8. 

\subsection{Selecting $D_E$}
Once $\tau$ has been selected, the estimation of $D_E$ is made by systematically asking when neighbors in dimension $D$ for $\S(n)$ remain neighbors when $\S(n)$ is expressed in dimension $D+1$. This method of false nearest neighbors reveals a global property of the source of the time series; namely, the minimum dimension $D_E$ within which the vectors $\S(n)$ can represent trajectories that contain no neighbors arriving through projection  from higher dimensions. The analysis of these global false nearest neighbors $FNN(D)$ is shown in the {\bf Right Panel} of Fig. (\ref{lor96d5amidt05}) where $\tau = 7$ is used. It is good to check that our results are robust against selecting $\tau = 7$; using $\tau = 6$ or $\tau = 8$ each yield $D_E = 5$. This is discussed in~\cite{abar96} in more detail.

Using $D_E = 5$, we can evaluate the Lyapunov exponents of the dynamical system at the source of the data $s(n)$. The methods for doing this are described in~\cite{abar96,kantz04}. Briefly summarized: one uses the development of trajectories in the $D_E$-dimensional space, and following a few trajectories nearby each other in this space from one location in space to another construct a local map from one set of points to the location where they go in one step. This permits one to read off the local Jacobian matrix step by step through out the trajectories in $D_E$-dimensional space. According to the Oseledec theorem~\cite{abar96} the sequential products of these $D_E \times D_E$ matrices when diagonalized yields the Lyapunov exponents. To accurately estimate all $D_E$ Lyapunov exponents, one uses a recursive $\Q\R\,$ decomposition.

Following this protocol, we find two positive exponents. There is one zero exponent telling us that the source of the data is some (unknown) differential equation for five state variables. There are two negative exponents. The sum of the Lyapunov exponents is negative, and the associated information dimension of the attractor is about 4.4~\cite{abar96,kantz04}.

Until this point, we have withheld information about the method in which the data was generated for this experiment. We now reveal that it is a $D = 5$ Lorenz 96 model with forcing term $F = 8.15$. Information was kept from the reader in order to illustrate that in order to perform such an analysis, information about the original system is not required.

Now that we have established estimates for $\tau$ and $D_E$, we want to use the observed time evolution $\S(n) \to \S(n+1)$ to train a network to implement this discrete time map in $D_E$-dimensions.

\section{Training a Selected Neural Network}

The idea is now to work in $D_E = 5$ dimensional space on vectors $\S(n)=[s(n),s(n+\tau),s(n+2\tau),...,s(n+ (D_E - 1)\tau)]$, and build a machine that learns the discrete time mapping $\S(n) \to \S(n+1)$.

To the scalar data  $\s(n)$ we add noise of mean zero and rms error $\sigma$ to form noisy scalar data $y(n) = s(n) + \mbox{Noise}(0,\sigma)$. In our calculations we selected $\sigma$ to be 2\% of the dynamical range of the observed data. With higher noise levels, the information content of each individual training pair is reduced, leading to a need of additional training pairs to achieve similar prediction capabilities.

Using these noisy scalar data we form a data library of many input/output $D_E$-dimensional vector pairs to be used at the input port at layer $l_0$ and the output port at layer $l_F$, to train a neural network. We use $k = 1,2,...,M$ members of this library as our training set.
\begin{eqnarray}
&&\Y^{(k)}(l_0) = \{y(k),y(k + \tau),y(k+2\tau),y(k+3\tau), y(k+4\tau)\}, \\
&&\Y^{(k)}(l_F) = \{y(k+1),y(k  + 1 + \tau),y(k + 1 + 2\tau),y(k + 1 +3 \tau), y(k + 1 + 4 \tau)\}, \nonumber
\end{eqnarray}
$k=1,2,...,M$ here and $D_E = 5$.

The network we choose is a Multi-layer Perceptron, and we wish to train it to take at the input vectors $\Y^{(k)}(l_0;n)$ and produce at the output vectors $\Y^{(k)}(l_F;n)$. At the input layer $l_0$ we have {\bf one} input port with $D_E$ slots. At the output layer $l_F$ we have {\bf one} output port with $D_E$ slots. The network has $l_F - 2$ hidden layers $l = \{l_1,l_2,...,l_F-1\}.$ At the hidden layers we have $D_{hl}$ active units ('neurons') at layer $l$.

As a function of the three quantities $\{l_F, D_{hl},M\}$: $l_F$, the number of layers or the `depth' of the network; $D_{hl}$, the number of active units in layer $l$ or the breadth of the network; and $M$, the number of distinct input/output pairs containing the information presented to the network for training, we wish to analyze, using statistical Physics methods, the quality of the training, the accuracy of the operation of the trained network on input/output pairs {\bf not} used in training, and the ability of the trained network to represent the information in the M data pairs. In the networks we develop here, we take $D_{hl}$ to be independent of $l$.

\subsection{The Action}

In much of machine learning one seeks to minimize a cost function evaluated at the input and the output layers of a selected network. We call the activity variables (`neurons') at layer $l$ $x_q(l)$ for active unit $q = 1,2, ..., D_{hl}$ in layer $l$. The cost function for each input/output pair is at time k
\begin{eqnarray}
    C(k) &=&
    \frac{R_m}{2}\frac{1}{D_{h0}+D_{hf}} \biggl[\sum_{q=1}^{D_{h0}}\,(x^{(k)}_q(l_0) - y(k + (q-1)\tau))^2 \nonumber \\
    &+& \sum_{q=1}^{D_{hF}}\,(x^{(k)}_q(l_F) - y(k + 1 + (q-1)\tau))^2\biggl],
\end{eqnarray}
where the noise or errors in the input and output data have been taken to be Gaussian with zero mean and diagonal precision matrix $R_m$. $D_{h0} = D_{hF} = D_E$ for us.

This is to be minimized subject to a layer-to-layer connection rule
\begin{equation} 
    x_q(l+1) = f_q \biggl(\sum_{v=1}^{D_{hl}}\W(l)_{qv}x_v(l)\biggr)\;\; q=1,...D_{h(l+1)},
\label{layerrule}
\end{equation}
with $\W(l)$ a matrix of weights to be determined in the minimization of $C(k)$.

If Gaussian errors with precision matrix $R_f$ are accepted in the layer-to-layer rule Eq. (\ref{layerrule}), then the full cost function is
\begin{equation}
    A(x(l);k) = C(k) + \frac{R_f}{2} \frac{1}{\sum_{l=l_1}^{l_F} D_{hl}}\sum_{l = l_0}^{l=l_F} \sum_{q=1}^{D_{h(l+1)}} \biggl( x^{(k)}_q(l+1) -  f_q(\sum_{v=1}^{D_{hl}}\W(l)_{qv}x^{(k)}_v(l))\biggr)^2, \nonumber
\end{equation} 
and we call this the `action', after its usage in statistical Physics, for a single input/output data pair chosen at time $k$.

When we have many input/output pairs, we add a label to the active states in the network $x_q(l) \to x^{(k)}_q(l)$, and our goal is to minimize
the action
\begin{eqnarray}
    A(x^{(k)}_q(l),\W(l)) &&= \frac{1}{M} \sum_{k=1}^M \biggl\{\frac{R_m}{2}\frac{1}{D_{h0}+D_{hf}} \biggl[\sum_{q=1}^{D_{h0}}\,(x^{(k)}_q(l_0) - y(k + (q-1)\tau))^2 \nonumber \\
    &&+ \sum_{q=1}^{D_{hF}}\,(x^{(k)}_q(l_F) - y(k + 1 + (q-1)\tau))^2\biggl] \label{totalaction} \\
    &&+ \frac{R_f}{2} \frac{1}{\sum_{l=l_1}^{l_F} D_{hl}}\sum_{l = l_0}^{l_F}\sum_{q=1}^{D_{h(l+1)}} \biggl( x^{(k)}_q(l+1) -  f_q\biggl[\sum_{v=1}^{D_{hl}}\W(l)_{qv}x^{(k)}_v(l)\biggr]\biggr)^2\biggr\}, \nonumber
\end{eqnarray}
with respect to the connection weight matrices $\W(l)$ and the activities $x^{(k)}_q(l)$. Minimizing this action recognizes that for each input/output training pair, the activity of the network nodes may differ, but averaging over all M presentations of pairs from the library will train a possible generalizable network characterized by the $\W(l)$ and any other fixed parameters in the nonlinear functions $f_q$.


\section{Use of the Action}

The action $A(\X)$, where $\X$ is the collection of all $x^{(k)}_q(l)$ in the network as well as the $\W(l)$ and other fixed parameters, is proportional to the negative of the logarithm of the conditional probability of the full state $\X$ conditioned on the M members of the input/output library, collected into a quantity $\Y$, used in the training set: $P(\X|\Y) \propto \exp[-A(\X)]$. An important use of this conditional probability density is the evaluation of expected values of functions $G(\X)$ on the variables $\X$, and this is evaluated by doing the integral
\begin{equation}
E[G(\X)|\Y] = \langle G(\X) \rangle = \frac{\int  d\X \, G(\X) \exp[-A(\X)]}{\int  d\X \, \exp[-A(\X)]}
\label{expect}
\end{equation}
It is here that the connection of machine learning with statistical Physics becomes apparent.

Estimating this integral can always be done with various Monte Carlo methods, and depending on the action $A(\X)$ surfaces in $\X$ may be accomplished by finding the maxima of $P(\X|\Y)$, or equivalently the minima of $A(\X)$. The latter method~\cite{laplaceold,laplacenew} is why we are interested in the paths $\X$ which yield minima of $A(\X)$. 

\section{Precision Annealing} 

We have developed a precision annealing (PA) approach~\cite{ye2014precision,ye2015physrev} for the minimization of the action Eq. (\ref{totalaction}) directed to finding the path with the smallest value of the action. The problem of finding the global minimum of the action, a nonlinear objective function of $\X$, is NP-complete~\cite{murty87}. PA is a continuation method~ \cite{allgower} in $R_f$ that begins at very small $R_f$ where the global minimum is a solution to minimizing a quadratic form; this can be done in a straightforward manner, and moves adiabatically in $R_f$ to quite high values. Formally as $R_f \to \infty$, the layer-to-layer rule used in constructing the network becomes precise and deterministic. 

While we have no mathematical proof that the global minimum is found, our numerical results indicate this may be the case. The PA method produces a set of minima of the action giving a numerical clue as to the roughness of the surface in path $\X$ space. It also finds low magnitude action minima with much higher rates of success than starting directly with large $R_f$.

The action surface $A(\X)$ depends, among other items, on the number of measurement pairs $M$, on the hyper-parameter $R_f$, and on the number of model layers between $l_0$ and $l_F$. As the number of hidden layers increases, the model architecture deepens.

\bigskip

\begin{algorithm}[H]
$R_{f} << 1$ \;
Choose $\X^0$ from a uniform distribution for each $N_I$\;
    \While{no individual $A(X')_{N_I}$ is substantially less than the group of $A$}{
        \ForEach{$N_I$}{
            Minimize $A(\X)$ using $\X^0$ as an initial guess\;
            Arrive at $\X'$\;
            $\X^0 = \X'$\;
            Update $R_f = R_f * \alpha$\;
        }
    }
    \caption{Precision() annealing algorithm}
\end{algorithm}
\bigskip

At the first step of PA we choose a solution to the optimization problem at $R_f = 0$ and select the states at the hidden layers as drawn from a uniform distribution with ranges known from the dynamical range of the input/output state variables. One can learn that dynamical range well enough by solving the underlying model forward for various initial conditions. We make this draw $N_I$ times, and now have $N_I$ paths $\X^0$ as candidates for the PA procedure.

Now we select a small value for $R_f$, call it $R_{f0}$, and use the previous $N_I$ paths $\X^0$ as $N_I$ initial choices in our minimization algorithm. After using that minimization procedure we find $N_I$ new paths $\X^1$ for the minimization problem with $R_f = R_{f0}$. This gives us $N_I$ values of the action $A(\X^1)$ associated with the new paths $\X^1$.

Next we increase the value of $R_f$ to $R_f = R_{f0}\alpha$ where $\alpha > 1$.  For this new value of $R_f$, we perform the minimization of the action starting with the $N_I$ initial paths $\X^1$ from the previous step to arrive at $N_I$ new paths $\X^2$. Evaluating the action on these paths $A(\X^2)$ now gives us an ordered set of actions that are no longer as degenerate. Many of the paths $\X^2$ may give the same numerical value of the action. However, typically the `degeneracy' lies within the noise level of the data $\approx (1/\sqrt{R_m})$. 

This procedure is continued until $R_f$ is `large enough' which is indicated by at least one of the action levels becoming substantially independent of $R_f$ and typically smaller than the others. 

Effectively PA starts with a problem ($R_f = 0$) where the global minimum is apparent and systematically tracks it and many other paths through increases in $R_f$. In doing the `tracking' of the global minimum, one must check that the selected value of $\alpha$ is not too large lest one leave the global minimum and land in another minimum. Checking the result using smaller $\alpha$ is always worthwhile.

It is important to note that simply starting with a large value of $R_f$, $R_f \approx 1$ or larger, places one in the undesirable situation of the action $A(\X)$ having multiple local minima into which any optimization procedure is quite likely to fall.

In the dynamical problems we have examined, one typically finds that as the number of measurement pairs $M$ is increased, more terms are added in the sum in Equation \ref{totalaction}, thus raising the action levels of minima disproportionately until there is one dominant minimum. This we attribute to the additional information from the augmented set of measurement pairs.

\subsection{Smallest Minimum; Not Necessarily a Convex Action}

As our goal is to provide accurate estimations of the conditional expected value of functions $G(\X)$ Eq. (\ref{expect}) where $\X$, a path in model space, is distributed  as $\exp[-A(\X)]$, we actually do not require convexity of $A(\X)$ as a function in path space. From the point of view of accurately estimating expected values, it is sufficient that the lowest action level be {\bf much} smaller than the second lowest action level. If the action value at the lowest level $A(\X_{\mbox{lowest}})$ is much smaller than the action value at the next minimum $A(\X_{\mbox{second lowest}})$, then by a factor $\exp[-\{A(\X_{\mbox{lowest}}) - A(\X_{\mbox{second lowest}})\}]$, the lowest path $\X_{\mbox{lowest}}$ dominates the integral to be done and provides a sensible choice for the path at which to evaluate the integral. 

We will see in the examples below that when the PA procedure is used we may encounter situations where the action is apparently not convex. However, it may have a distinct smallest action level, much smaller in magnitude than the next lowest action level. That lowest level is expected to give a path which gives an accurate estimation to the expected value of functions \small $G(\X)$ \normalsize. This may occur in cases where sufficient information from the data has been transferred to the model, and this can indicate the size model adequate for the problem posed. 

\section{Action Levels for Our Time Series $\{s(n)\}$}

We will now build and train a feedforward MLP~\cite{prvaranneal18} to learn the function $s(n) \to s(n+1)$ using $D_E$-dimensional data pairs from our library. We examined networks with $D_E$ = 5 dimensional input ($l_0$) and output ($l_F$) layers and 1-5 hidden layers each with the same number $D_h$ of active units (`neurons'). The nonlinear function operating from layer-to-layer was chosen to be $\tanh(\bullet)$. We use the Python based program VarrAnneal~\cite{prvaranneal18} to perform the minimization of the action at each value of $R_f/R_m > 0$.

To prepare our data for these network choices, we first scaled all of the noisy inputs $y(n)$ to lie within the range $[-1,1]$ via
\begin{equation}
    y(n) \to \frac{2 y(n) - (y_{max} + y_{min})}{y_{max} - y_{min}},
\end{equation} 
where $y_{max,min}$ are the maximum and minimum values taken by the noisy data. These scaled values were used to construct our data library of input/output pairs.

\subsection{Two Hidden Layers; $D_h$ = 15; M = 50,..., 1200}

Using PA and systematically moving $R_f/R_m$ from $R_{f0}/R_m \approx$ 10$^{-8}$ to $R_f/R_m \approx$ 10$^{11}$ we evaluated $A(\X)$ for $M = 50, 100, ..., 1200$ input/output pairs. $R_f$ was slowly increased using $\alpha = 1.1$ At each value of $R_f/R_m$ we used $N_I = 20$ initializations of the PA procedure.

\begin{figure}[tbph] 
  \centering
 \includegraphics[trim={200 50 400 100}, clip, width=0.45\textwidth]{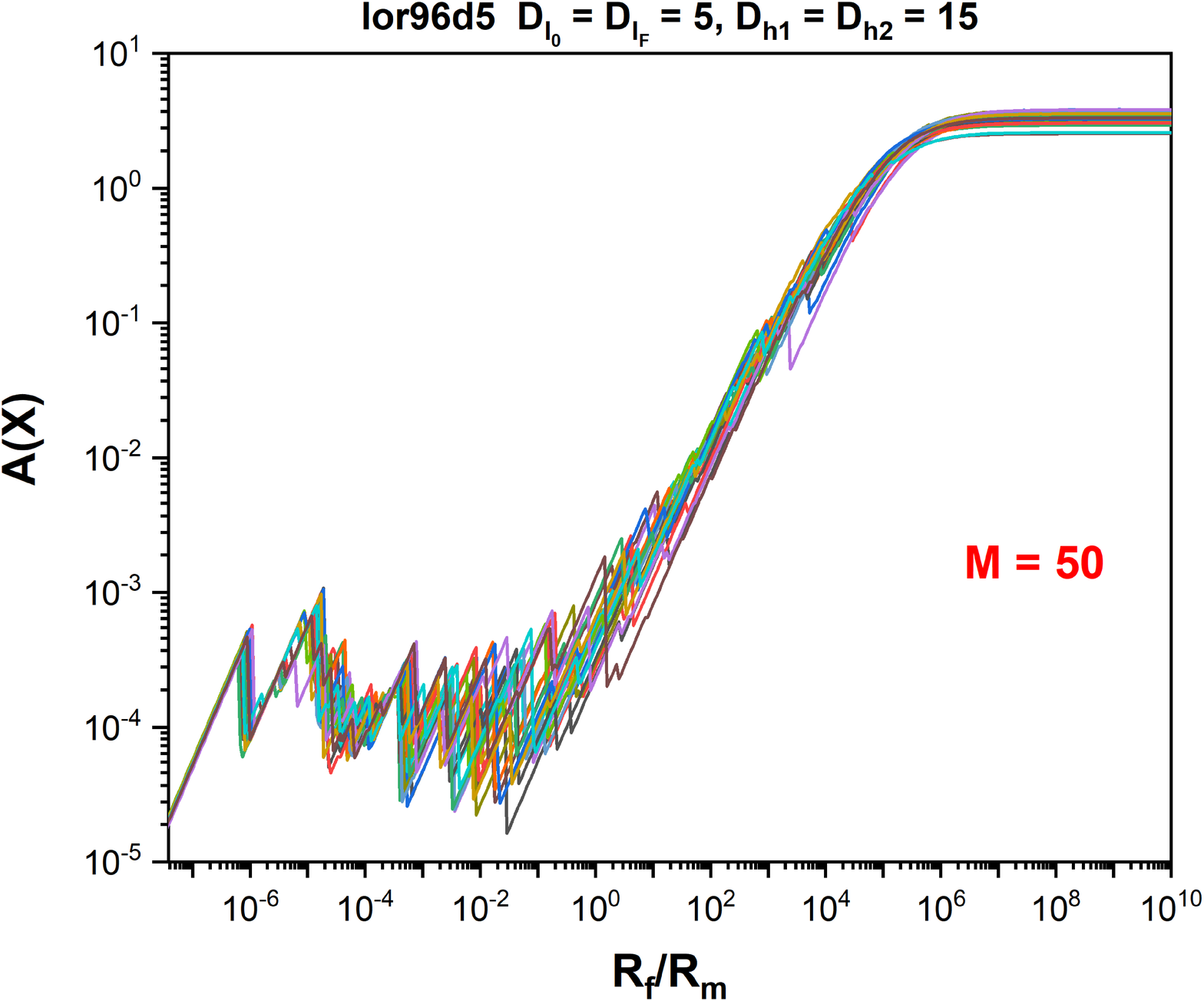}
 \hspace{10pt}
 \includegraphics[trim={200 50 400 100}, clip, width=0.45\textwidth]{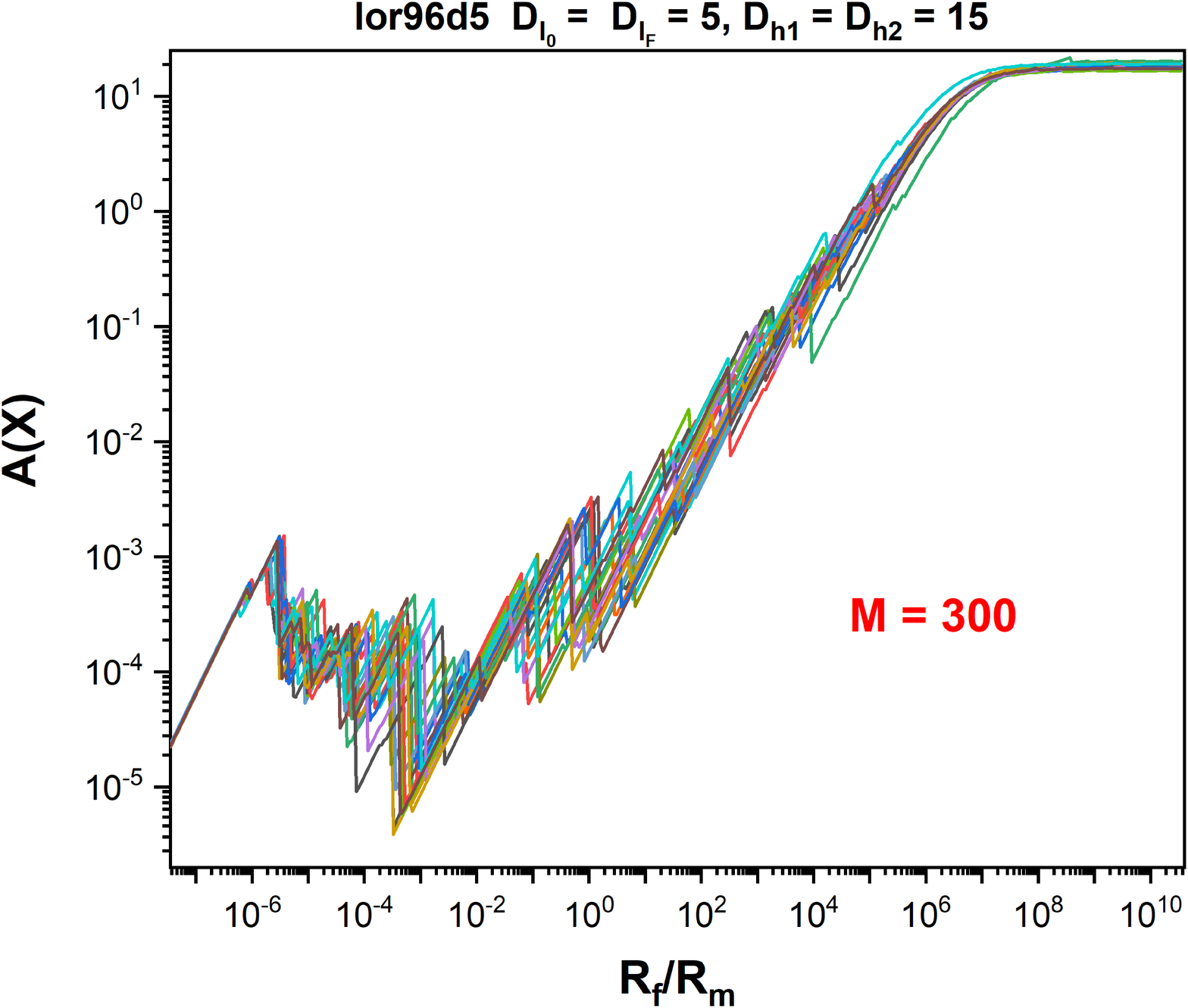}
 \includegraphics[trim={200 50 400 100}, clip, width=0.45\textwidth]{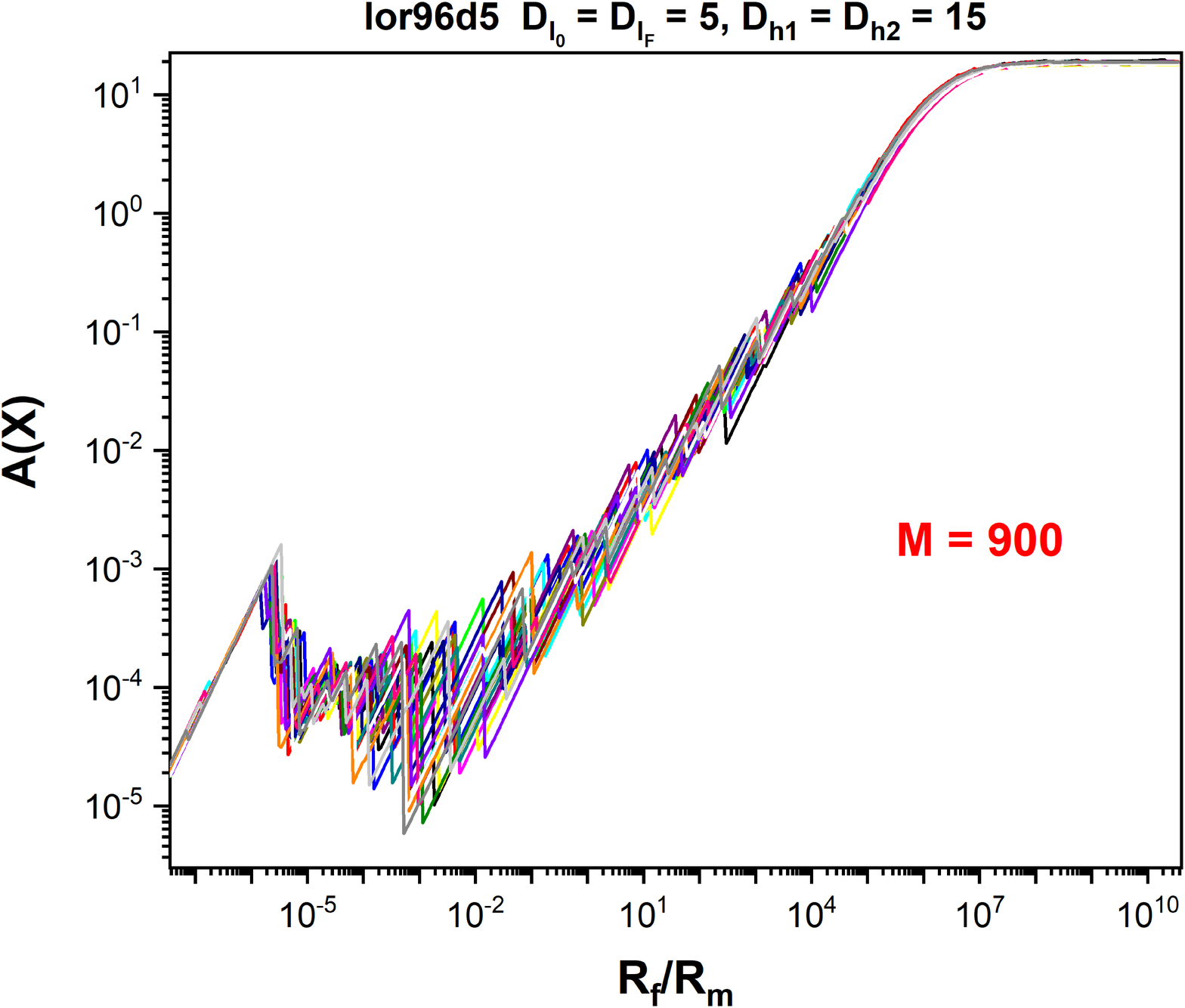} 
 \hspace{10pt}
 \includegraphics[trim={200 50 400 100}, clip, width=0.45\textwidth]{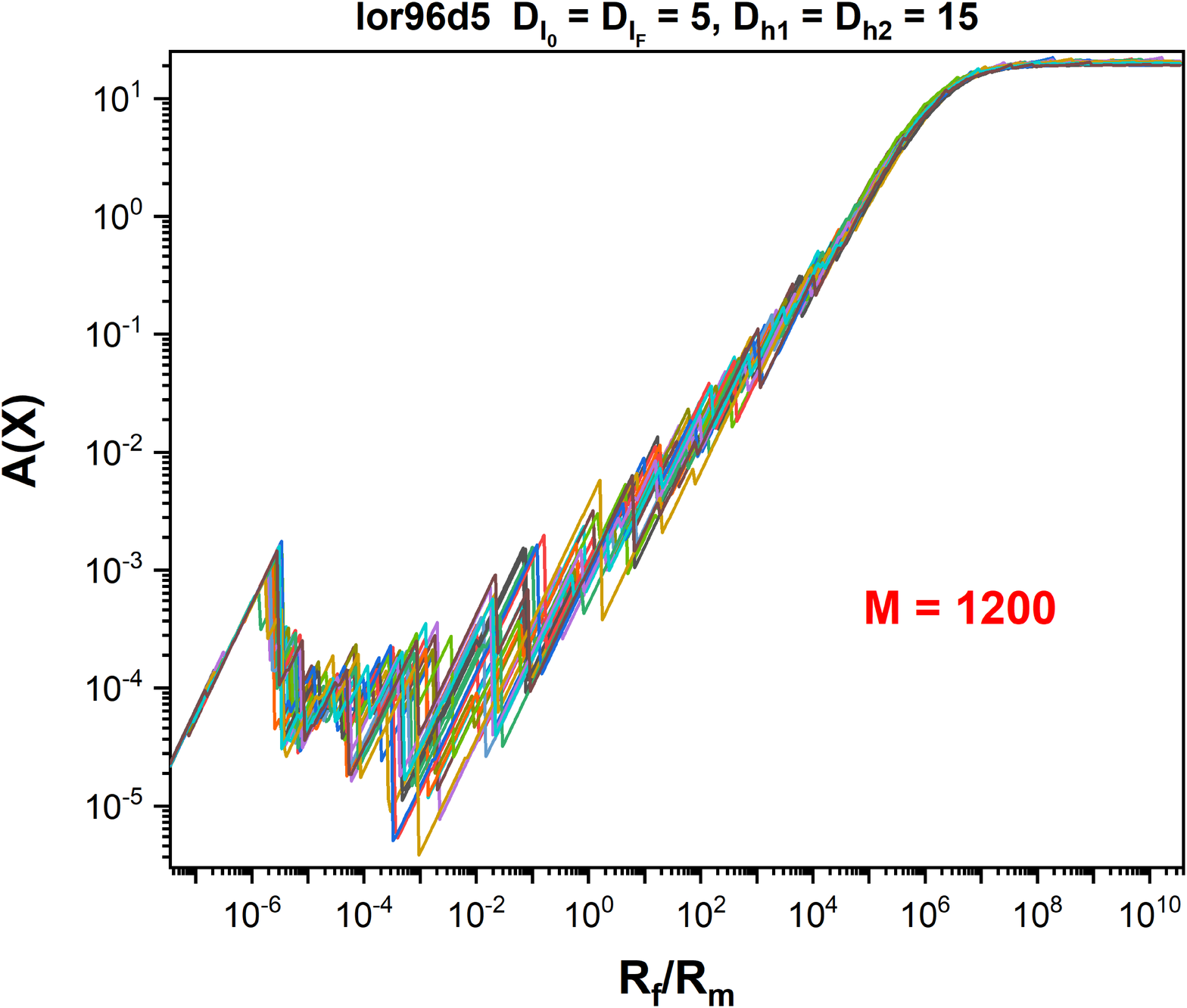}
 \caption{Action Levels as a function of $R_f/R_m$ for the time series data $s(n)$ input into our two hidden layer MLP as a $D_E$ dimensional data vector. The number of I/O pairs for these calculations were $M = 50, 300, 900, 1200$. $N_I = 20$ action levels associated with the $N_I$ initializations of the optimization algorithm used at each value of $R_f/R_m$. In these calculations $\alpha = 1.1$ in the PA procedure. Note that as M increases the action level for large $R_f/R_m$ rises and then saturates becoming effectively independent of $R_f/R_m$. This is presented more 
precisely in Fig. (\ref{maxAvsM}).} 
  \label{lor96d5_5_15_15_5M50}
\end{figure}

\begin{figure}[tbph] 
  \centering
  \includegraphics[width=5.67in,height=3.96in,keepaspectratio]{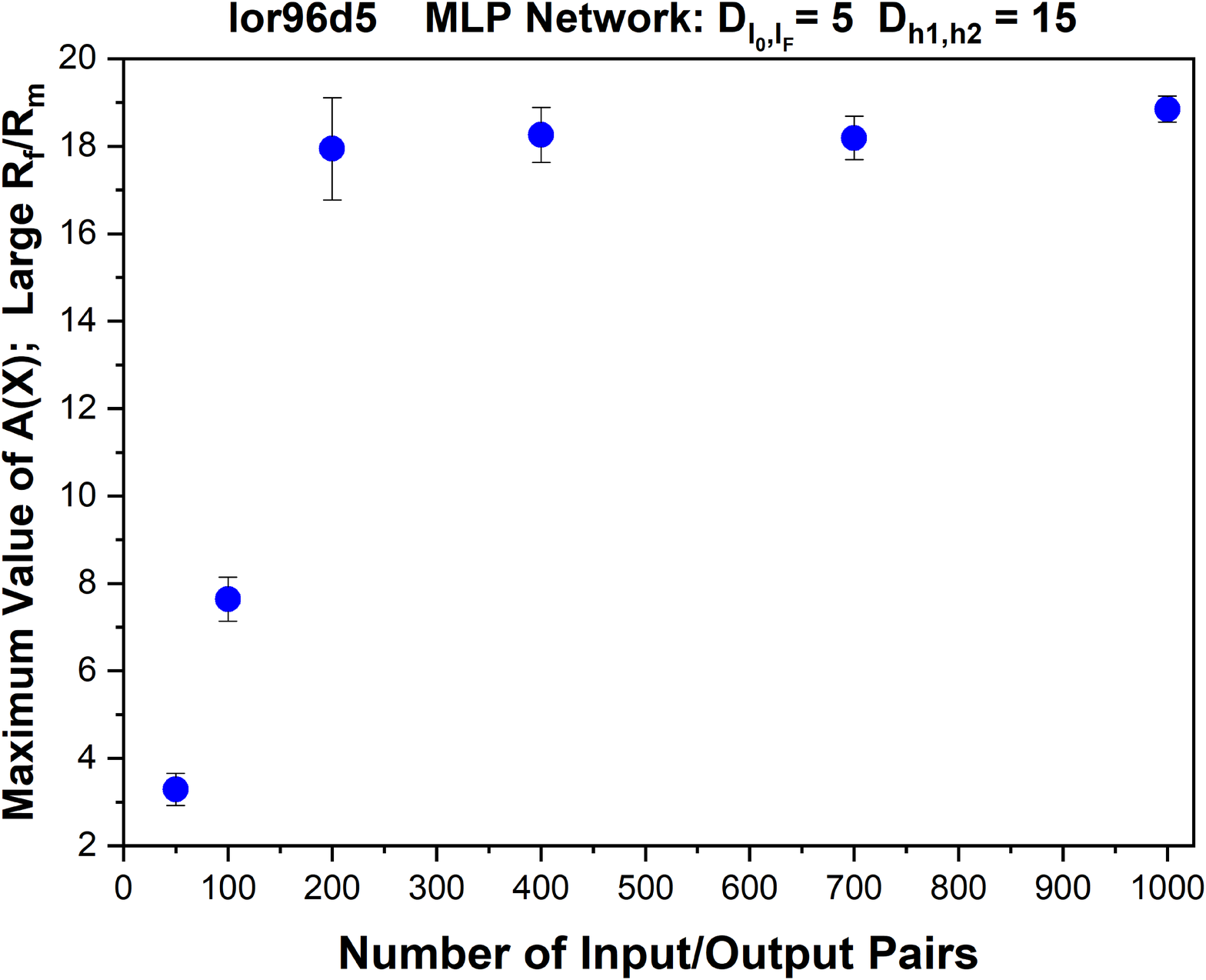}
  
  \caption{Display of the average and standard deviation of the $N_I = 20$ largest action levels versus the number $M$ of I/O pairs in the training set for our time series $\{s(n)\}$. As seen in the action levels plots as $M$ increases the maximum action levels grow then saturate as the network reaches a full representation of the information in the data pairs. This kind of calculation allows the network designer to determine for a given network architecture how many samples from the I/O library of pairs will be needed to fully train the network. The expected $A(\X)$ saturates near $M$ = 200.}
  \label{maxAvsM}
\end{figure}

We first examine the structure of the action levels as a function of $R_f/R_m$ for $M = 50, 300, 900, 1200$ I/O pairs. This is displayed in Fig. (\ref{lor96d5_5_15_15_5M50}). Note that the action levels become nearly independent of $R_f/R_m$ for large values of this hyperparameter. Equally interestingly is the initial rise of $A(\X)$ for large $R_f/R_m$ as $M$ increases. Then  this saturates as the information in the time series $s(n)$ is represented fully in the network. See Fig. (\ref{maxAvsM}). Recall we call this the information content as, up to a constant $\langle A(\X) \rangle = \langle - \log[P(\X)] \rangle$.

\section{Errors in Training and Validation}

Once we have trained the proposed network, we can evaluate its quality when performing the task we have set it. In the example we have discussed here, that task in summarized as: when presented with a $D_E$-dimensional vector of inputs, created from time delays of a signal $s(n)$, accurately produce  the next element of the time series $s(n+1)$. The quantities $s(n)$ and $s(n+1)$ are the first components of the data vectors. 

We have tested (or validated) the operation of the network both on the data used to train the network and on data held aside in our library of I/O pairs. The latter is often called the ``test'' set or validation set or prediction set portion of the total data available to us~\cite{Frank2001}.

The error on the training set as a function of the number $M$ of I/O pairs used to train the network is given as
\begin{equation}
    MSE_{\mbox{Training Error}}(M) = \frac{1}{M} \sum_{k=1}^M \frac{1}{D_E}\sum_{q=1}^{D_E} (x^{(k)}_q(l_F) - y(k+1 + (q-1)\tau))^2
\end{equation}

\begin{figure}[tbp] 
  \centering
  \includegraphics[width=5.67in,height=4.34in,keepaspectratio]{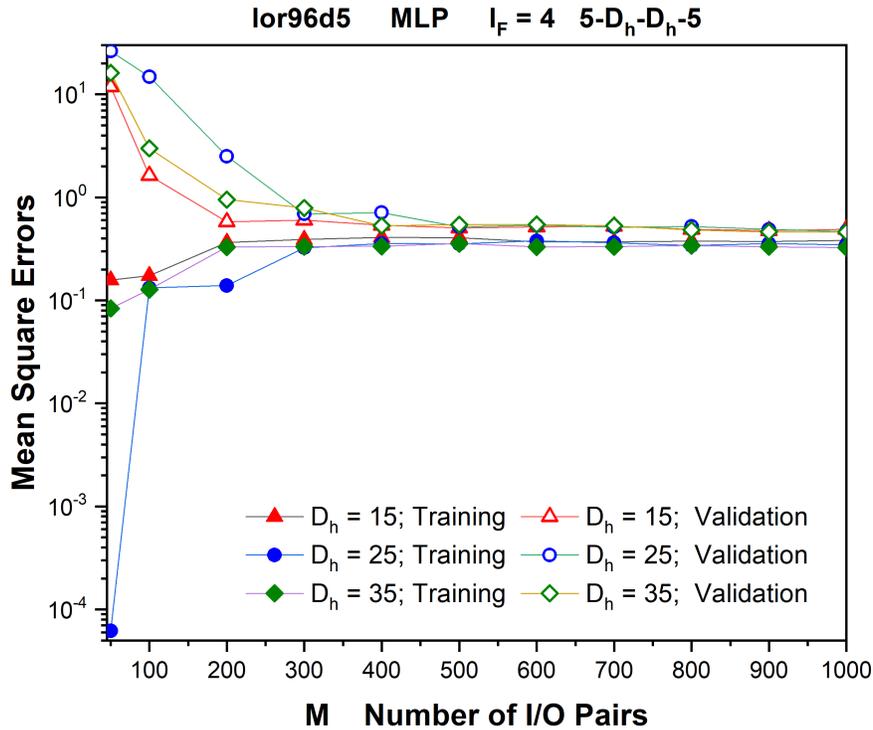}

  \caption{Mean Square Errors (MSE) in the representation of the information in our scalar time series $s(n)$ using a network with four layers. When $M$ is approximately 300, all of these errors become essentially independent of $M$ and very close to being equal. This tells us that to using this network of `sigmoids and wires' we need no more than a few hundred samples of the data to predict as well as we can. No new capability is revealed when $M$ grows beyond that level. This is consistent with the knowledge of the Lyapunov exponents determined for the time series presented to us~\protect\cite{abar96,kantz04}.}
  \label{lor96d5-5-Dh-Dh-5_Dh15-25-35errors}
\end{figure}

This compares, in a least squares sense, the $D_E$-dimensional output $x^{(k)}_q(l_F)$ from the trained network with the data from the {\bf training} set, $y(k+1 + (q-1)\tau)$, that are the output side of the input/output training pairs. The input to the trained network are the values $\Y^{(k)}(l_0) = \{y(k),y(k + \tau),y(k+2\tau),y(k+3\tau), y(k+4\tau)\}$; the trained network operates on this $D_E$-dimensional vectors producing the output $x^{(k)}_q(l_F);\;q=1,2,...,D_E$.
We plot this as a function of $M$, the number of input/output pairs used in the training procedure, and in Fig. (\ref{lor96d5-5-Dh-Dh-5_Dh15-25-35errors}) we also examine the dependence on the number of active units (``neurons'') in each of the two hidden layers in the network.

We also can determine the accuracy of the trained network when acting on inputs selected from I/O pairs {\em not} used in the training of the network. This `validation' error is evaluated as
\begin{equation}
    MSE_{\mbox{Validation Error}}(M) = \frac{1}{M_{\mbox{total}} - M} \sum_{k=M}^{M_{\mbox{total}}}\frac{1}{D_E} \sum_{q=1}^{D_E} (x^{(k)}_q(l_F) - y(k+1 + (q-1)\tau))^2.
\end{equation}

This compares the $D_E$-dimensional output $x^{(k)}_q(l_F)$ from the trained network with the data from the set of input/output pairs that {\bf were not used} during the training, $y(k+1 + (q-1)\tau)$, that are the output side of the input/output pairs from the data library. We plot this as a function of $M$, the number of input/output pairs used in the training procedure. All of the I/O pairs from the data library not used in training were used in this validation error estimate. $M_{\mbox{Total}} = 84971 \gg M$.

In the input layer there are five ports into which a  vector $\Y(n) = [y(n), y(n+\tau), y(n+2\tau), ..., y(n+4\tau)]$ is presented. In each of the hidden layers there are 15 or 25 or 35 active units (`neurons'). At the output layer there are five ports within which a vector $\Y(n+1) = [y(n+1), y(n+\tau+1), y(n+2\tau+1), ..., y(n+4\tau+1)]$ is estimated. In Fig. (\ref{lor96d5-5-Dh-Dh-5_Dh15-25-35errors}) we show the MSEs in the estimation/training window and the prediction/generalization window as a function of the number of active units $D_h$ in the hidden layers and as a function of the number of input/output pairs $M$ used in training the network.
The results show that for small $M$ the training and validation errors differ substantially, but as $M$ increases, enough information lies in the training set of $M$ training I/O pairs that the overall training error levels out when the network has completed its representation of the information in the data. Similarly, while the validation error is large for small $M$, as the network becomes `well trained' (represents the information in the data series) the prediction MSE is essentially the same as the MSE in training. This result is consistent with the observation that the maximum value of the action levels for large $R_f/R_m$ becomes independent of $M$; see Fig.(\ref{maxAvsM}). 

Fig. (\ref{lor96d5est_prederrorsHL2,3,4vsM}) examines the training and validation MSEs as a function of the number of layers in the network. The number of hidden layers is $l_F -2$, and we have evaluated this, using our achitecture, for $l_F =\,$ 4,5, and 6.

\begin{figure}[htbp] 
  \centering
  \includegraphics[width=5.67in,height=4.34in,keepaspectratio]{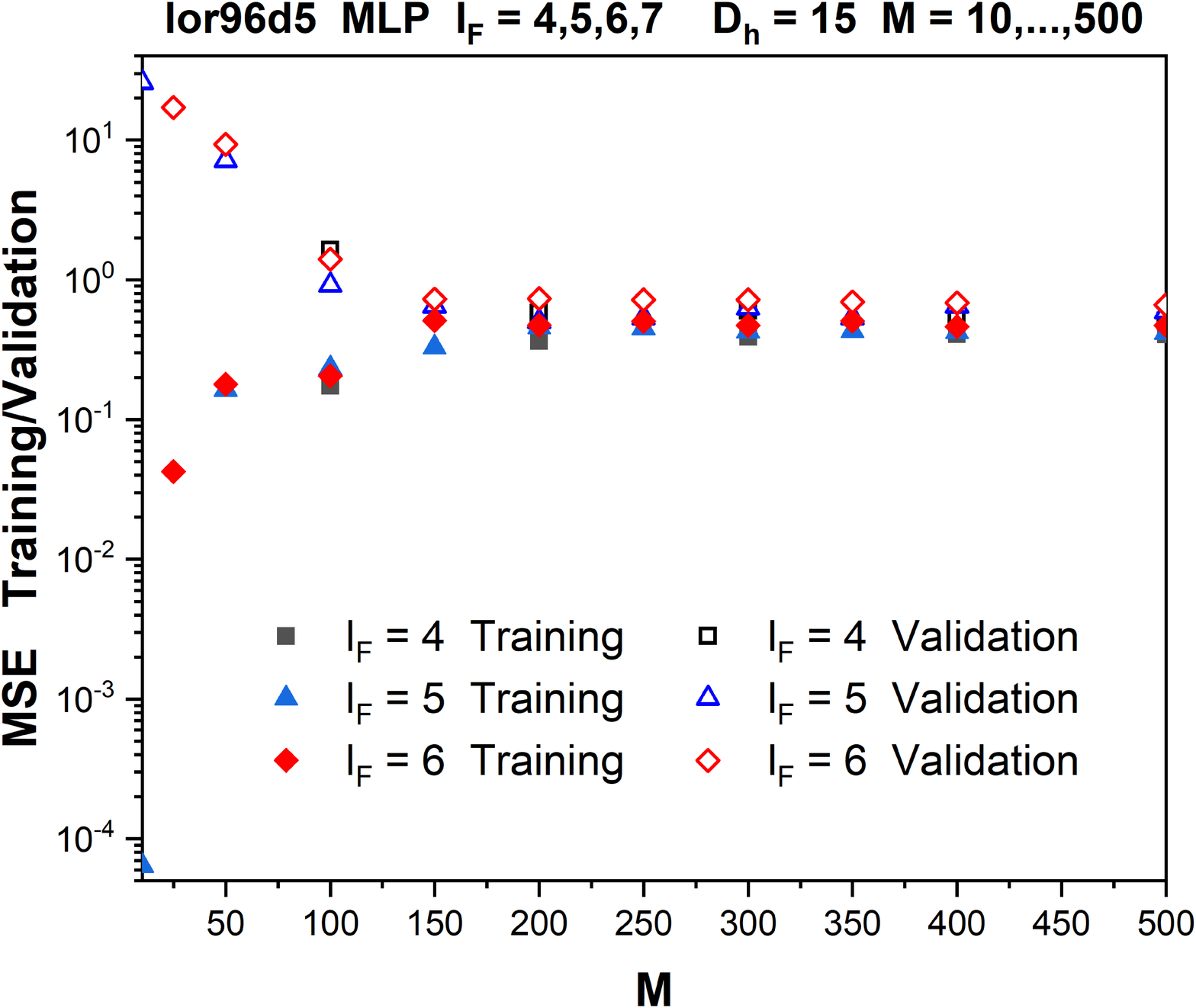}

  \caption{The Mean Square Errors (MSEs) in training and validation (estimation and prediction) for the network architecture  with $l_F = 4, 5,6$ layers. The input and output layers have five ports as before. All hidden layers have 15 active units. We see that when $M \ge 200$ or so, the performance of the network architecture becomes independent of the number of training samples as well as of the number of layers in the network. This result, as in the numerical data displayed in Fig. (\ref{lor96d5-5-Dh-Dh-5_Dh15-25-35errors}), shows how the PA method can capture the essential information processing power of a selected architecture of a multi-layer perceptron. More to the point, it informs us how many $M$ input/output pairs are required to perform the task set to the machine.}
  \label{lor96d5est_prederrorsHL2,3,4vsM}
\end{figure}

\begin{figure}[htbp] 
  \centering
  \includegraphics[width=5.67in,height=4.34in,keepaspectratio]{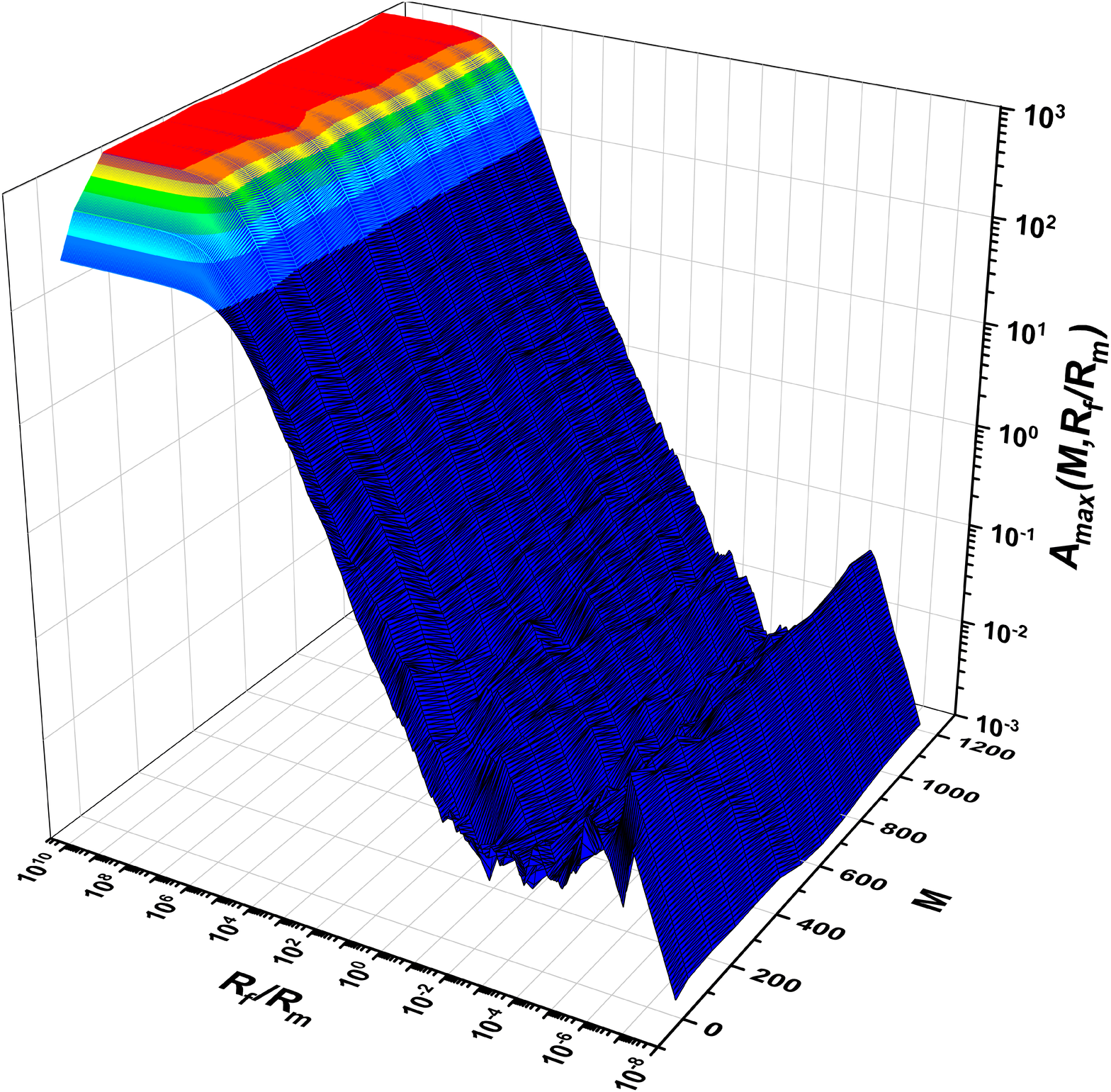}

  \caption{A three dimensional plot of the action for the $l_F = 4;\,D_h = 15$ network as a function of $R_f/R_m$ and the number of input/output pairs $M$ presented for training. It is clear here that a distinct plateau appears in the action, to be thought of as the information content $(-\log[P])$ within the time series data now represented in the network. }
  \label{amax3DMRf_Rm}
\end{figure}

We display the dependence of the action on the number of input/output samples $M$ and $R_f/R_m$ relevant in the PA algorithm in Fig. (\ref{amax3DMRf_Rm}) to further illustrate the outcome of our MLP network instantiation.

\begin{figure}[tbp] 
  \centering
  \includegraphics[width=5.67in,height=4.33in,keepaspectratio]{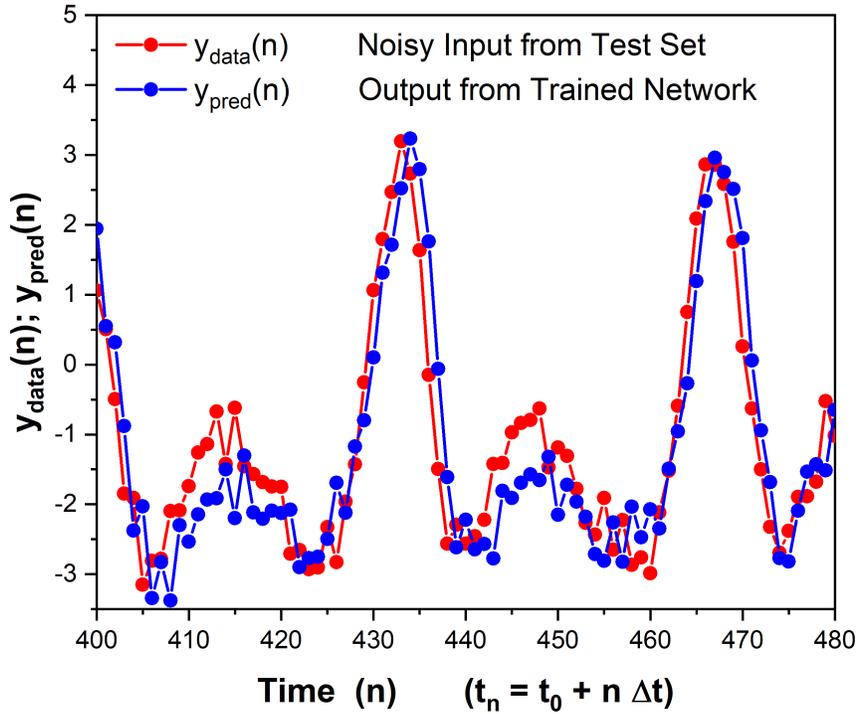}
    \caption{Using the $l_F = 4;\;D_{in} = D_{out} = 5; D_{h} = 15$ network, we show how, after training with 400 pairs of 5 dimensional noisy inputs $\{y(n),...,y(n+4\tau)\}$ and 5 dimensional outputs $\{y(n+1),...,y(n+4\tau+1)\}$, this network is able to predict one step ahead for a new five dimensional input. }
  \label{lor96d5_yred_n_ydata_ntestset}
\end{figure}

In Fig. (\ref{lor96d5_yred_n_ydata_ntestset}) we display the predictions produced by our trained MLP network after the training using $M = 400$ input/output pairs of segments of the noisy time series starting at $y(n-1)$ and predicting $y(n)$ in comparison with the known value of $y_{data}(n)$. The training is performed in $D_E$-dimensional space and the the output of the network is also in $D_E$ dimensional space. We display only the first component of the $D_E = 5$ dimensional proxy state space vector as that is our (noisy) measured quantity. The predictions are what this network has been trained to do.

\begin{figure}[tbp] 
  \centering
  \includegraphics[width=5.67in,height=4.34in,keepaspectratio]{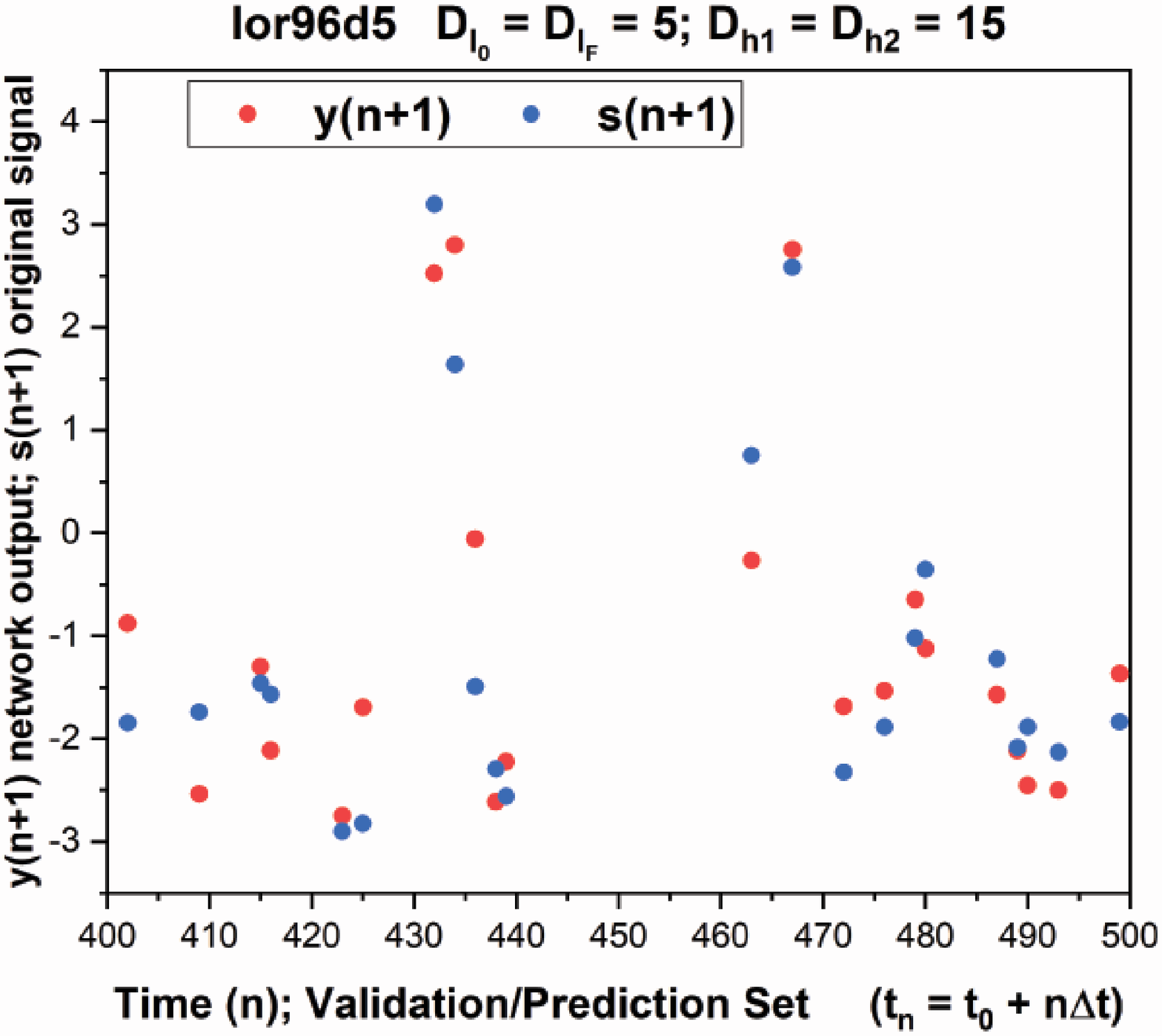}

  \caption{Now we take the trained $l_F = 4;\;D_{in} = D_{out} = 5; D_{h} = 15$ network and regard it as a dynamical system taking $s(n) \to s(n+1)$. When asked to predict as a dynamical system, namely without continuing information input from the data, the network is not performing well. Of course,  the network was not trained to this task. }
  \label{lor96d5predasdynamics}
\end{figure}

If we ask another question of the network: take the trained network as a dynamical system, namely, train the network using $M=400$ input output pairs, then use the trained network to predict $y(n) \to y(n+1)$ forward from the training window, we find the results in Fig. (\ref{lor96d5predasdynamics}). On this task the trained network does not perform as well as on the task it was trained to do

\section{Summary and Discussion}

Using the interpretation of a familiar machine learning task: using the information flow in a scalar time series to train a rather standard multi-layer perceptron (MLP) to predict one step forward in the time series, as the equivalent of a statistical data assimilation (SDA) task~\cite{abar18}, we have shown that using the precision annealing~\cite{ye2014precision,ye2015physrev} training methods of SDA, given the model architecture, leads to a network whose action ($\propto -\log[P(\X)])$ rapidly becomes independent of the precision of the model as well as independent of the number $M$ of input/output model pairs and independent of the number of model layers for two or more hidden layers.

We attribute this independence to the class of models having captured the information content within the time series, and thus the method of training reveals how one may use precision annealing to estimate the number of input/output pairs required for excellent training and accurate prediction/generalization. Efforts to this end have been mainly curve-fitting learning curves, for example:~\cite{figueroa2012},~\cite{beleites2013}. Furthermore, as knowledge of the conditional expected values of model state variables is what we wish to utilize approximations of the conditional probability distribution of model states $P(\X|\Y)$ for, $\X$ is the collection of all model states at all layers, and $\Y$ is the collection of all input/output noisy data pairs, we can see how to properly limit the number of data pairs in a training set. This can be important in practical applications.

The data set used in these experiments was generated with the Lorenz96 model equations. Our analysis assumed no knowledge of this to illustrate that the decisions made for preparing the data can be made independent of its source. In curating the data we employed a technique from nonlinear time series analysis~\cite{abar96,kantz04} that, while well known in the analysis of time series from nonlinear sources, has been used only once~\cite{Frank2001}, as far as we could tell, in a machine learning context over some decades. Considering its utility, we employed it here in a bit of detail as a friendly suggestion for future time series investigations.

In our earlier paper~\cite{abar18} introducing the analogy between machine learning and SDA, we noted the saturation of actions and prediction quality in a less structured example. We have shown it again here with an attribution to its information theoretic origin. As precision annealing within a Lagrangian training approach from classical methods of variational 
principles~\cite{gfomin,marsdenwest,nirag17} is utilized by us, the success may also be attributed to the capability of precision annealing to follow 
the global minimum of the action even though it is nonlinear in its variables $\X$~\cite{murty87}. This is the value of the action that maximizes the contribution of the conditional expected values of many quantities of interest.

The training method for the MLP network follows that for variational principles in data assimilation~\cite{even,bocquet17,abar13,marsdenwest} and control theory~\cite{kirk,gfomin} in which `backpropagation' procedures are absent and the methodology is well organized and principled. An additional value of the methods used here and in these references is that the symplectic structure of the variational principles is maintained~\cite{gfomin,marsdenwest,nirag17,abar18}.


\section*{Acknowledgment} Participation by PJR in this research has been partially funded by Deutsche Forschungsgemeinschaft (DFG) through grant CRC 1294 "Data Assimilation" (project A06).

\newpage

\bibliography{NECO-02-19-3407-Bib}
\bibliographystyle{apacite}

\end{document}